\documentclass[letterpaper, 10 pt, conference]{ieeeconf}  

\IEEEoverridecommandlockouts                              

\usepackage{amsmath,amssymb}
\usepackage{cite}
\usepackage{multirow}
\usepackage{graphicx}
\usepackage{array}
\usepackage{booktabs}
\usepackage{float}
\usepackage[colorlinks=true, linkcolor=blue]{hyperref}
\usepackage{orcidlink}
\usepackage{cleveref}

\usepackage{array,tabularx,makecell}

\newcolumntype{Y}{>{\centering\arraybackslash}X}

\title{\LARGE \bf
MiniPose: Miniature-Based Real Data Collection for 6D Pose Estimation of Large Industrial Pipe Segments
}

\author{
Hyunjun Kim$^{1}$\orcidlink{0009-0005-6260-8059},
Youngjin Seo$^{1}$\orcidlink{0009-0004-8247-6340},
Sungwon Seo$^{2}$\orcidlink{0000-0001-7862-7936}, 
Jiyou Shin$^{2}$\orcidlink{0009-0007-8952-1644},\\
Kunwoo Kim$^{1}$\orcidlink{0009-0000-5794-414X},
Chanyong Park$^{2}$\orcidlink{0009-0005-6779-1489},
and Hyungpil Moon$^{1,2,*
}$\orcidlink{0000-0002-1091-0716}%
\thanks{This work was supported by the Technology Innovation Program 2410012091 (00443027, Development of a mobile manipulator robot system based on multi-collaboration for manipulating and assembling 500 kg large and heavy parts) funded by the Ministry of Trade, Industry \& Energy (MOTIE, Korea).}
\thanks{
1: The authors are with the Faculty of Intelligent Robotics, Sungkyunkwan University, Suwon 2066, South Korea
(e-mail: \mbox{hyunjun.1104@g.skku.edu}; \mbox{soh2879@g.skku.edu}; \mbox{kkwkmw@g.skku.edu}; \mbox{hyungpil@skku.edu})
}
\thanks{
2: The authors are with the Faculty of Mechanical Engineering, Sungkyunkwan University, Suwon 2066, South Korea
(e-mail: \mbox{danny384@g.skku.edu}; \mbox{ssw0536@g.skku.edu}; \mbox{pcy23107@g.skku.edu}; \mbox{hyungpil@skku.edu})
}
}
\newsavebox{\tabbox}
\begin{document}
\raggedbottom

\maketitle
\thispagestyle{empty}
\pagestyle{empty}

\begin{abstract}

6-DOF pose estimation of large and heavy industrial objects, such as large piping systems in shipyards and structural components in petrochemical plants, remains relatively underexplored in robotic perception.
Repeatedly repositioning large objects is physically and spatially demanding, making it difficult to acquire real training data with sufficient pose diversity.
Synthetic data offer a scalable alternative, but faithfully reproducing real RGB-D observations, particularly the structured measurement errors of real depth sensors, remains difficult.
We propose MiniPose, a practical pipeline for miniature-based real RGB-D data collection that addresses both the data-acquisition burden and the sim-to-real gap.
We fabricate a scaled-down replica and capture RGB-D data of it in diverse poses using a real RGB-D camera.
RGB images are used directly, while depth maps are converted to a geometrically consistent representation of the full-scale object via a simple inverse scaling transformation.
This allows training to incorporate real appearance and sensor measurement characteristics that are difficult to reproduce using synthetic data alone.
To examine whether the benefit depends on a particular network, we evaluate two RGB-D pose estimators with different formulations, PVN3D and RDPN6D.
Mixing at most 1,000 miniature images per pipe type, less than 10\% of the training data, with 10,000 synthetic images
improves the mean ADD-S AUC across pipe types over synthetic-only training for both estimators.
A comparison with training on full-scale real data further suggests that miniature data can serve as a proxy where full-scale collection is difficult, reducing the data-collection burden for large industrial objects.

\end{abstract}

\vspace{\baselineskip}

\section{INTRODUCTION}

Recent deep learning-based 6D pose estimation methods have achieved strong performance.
However, existing research has largely focused on small objects that can be directly handled by robotic manipulators~\cite{ordoumpozanis2025reviewing,gouda2025mr6d}.
In contrast, 6D pose estimation for physically large and heavy industrial components remains comparatively underexplored~\cite{gouda2025mr6d}.
Existing studies on such components use geometry-based vision measurement to guide assembly rather than learned pose estimators~\cite{cheng2022bridge,zhou2021gear}.
Small objects can be freely repositioned by hand or on a turntable to collect real training data from diverse poses.
Large objects cannot be moved this way due to their size and weight, which makes it difficult to acquire training data with sufficient pose diversity.
Camera placement is also more challenging because capturing the full shape of a large object requires positioning the camera farther away.

To address this data scarcity, many studies generate synthetic data from 3D CAD models and reduce the sim-to-real gap~\cite{tremblay2018dope,hinterstoisser2018synthetic}.
For RGB, domain randomization of lighting, backgrounds, and textures can reduce the appearance gap relatively easily~\cite{tobin2017domain}, whereas realistically reproducing the behavior of real depth sensors remains difficult~\cite{zhang2023active,dai2022dreds,planche2017depthsynth}.
Depth noise has been handled either by explicitly modeling it or by correcting the measurements, but both depend on the characteristics of a specific sensor~\cite{landau2016simulating,meister2013,liu2026manipulation}.
Consequently, a sim-to-real gap can remain between synthetic training data and real observations, degrading model generalization to real-world environments~\cite{planche2017depthsynth,dai2022dreds}.

\begin{figure*}[t]
    \centering
    \includegraphics[width=0.85\textwidth]{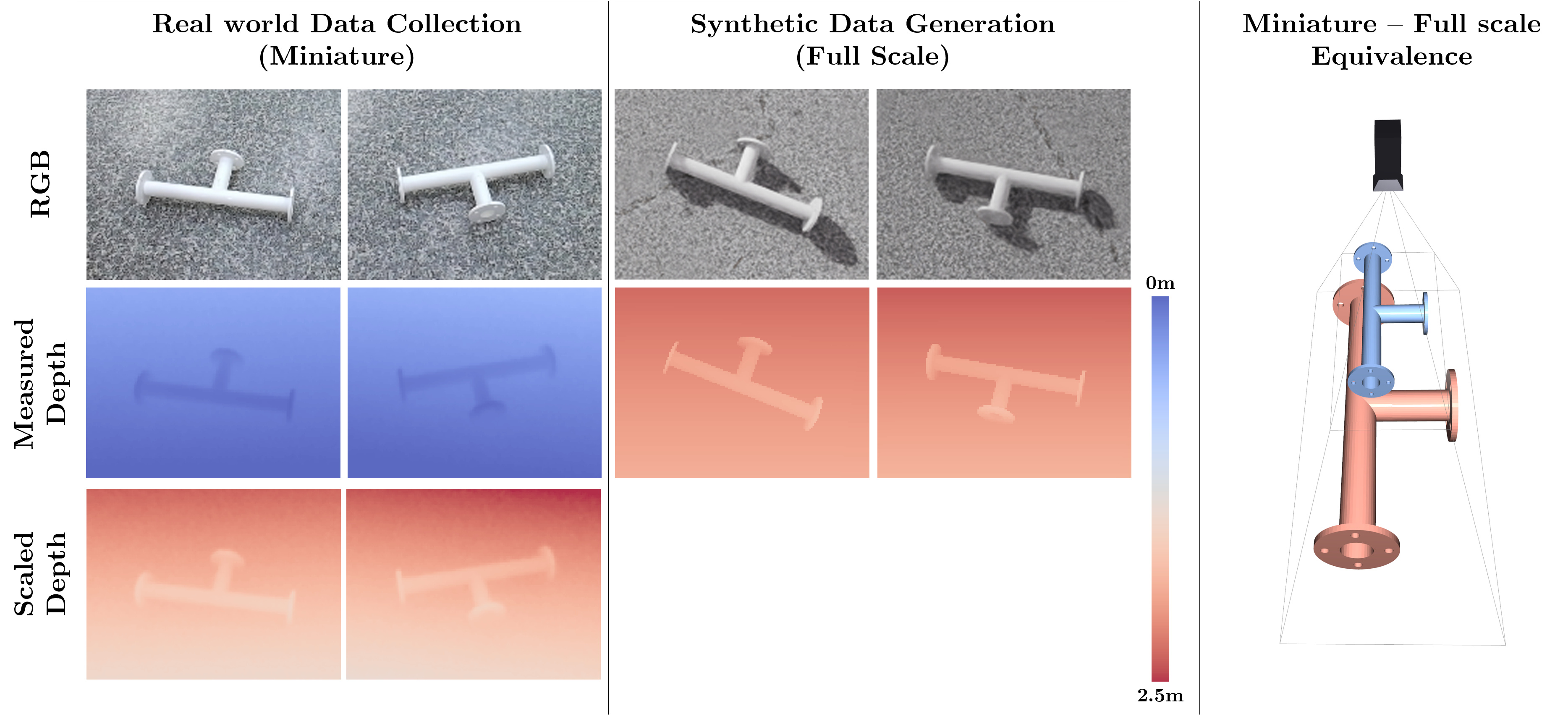}
    \caption{\textbf{Overview of the proposed approach.} For the miniature (left), the camera directly measures depth at the miniature's physical scale (about 1\,m, blue). Applying the depth scaling of~\Cref{sec:depth_scaling} rescales this to the full-scale range (about 2\,m, red). The full-scale synthetic renders (middle) provide depth directly in this range. The scale equivalence between the miniature and the full-scale object is illustrated (right).}
    \label{fig:pipeline}
\end{figure*}

Rather than further complicating the simulation model, we address both the data scarcity and the sim-to-real gap at the data-acquisition stage by proposing MiniPose, a miniature-based RGB-D data collection pipeline.
Scaled-down replicas of large objects are easy to reposition, enabling more efficient collection of real RGB-D data across diverse poses than with the full-scale objects.
They can also be captured at a shorter camera distance, reducing the space required for data acquisition.
Because a miniature is captured with a real RGB-D sensor, it provides both real RGB and real depth at once, without a noise model of the camera.
The captured data are adapted to the full-scale training domain through a simple geometric transformation and combined with synthetic data.
This provides a convenient way to incorporate real observations into training without repeatedly manipulating the full-scale object.

We validate the proposed method on six representative industrial pipe types commonly used in heavy-industry settings.
Mixing depth-scaled miniature images, which constitute less than 10\% of the training set, with synthetic images increases the average ADD-S AUC over synthetic-only training for both PVN3D~\cite{he2020pvn3d} and RDPN6D~\cite{hong2024rdpn6d}, and also outperforms a Time-of-Flight (ToF) noise model and test-time depth correction with the Camera Depth Model (CDM)~\cite{liu2026manipulation}.
A comparison with training on directly captured full-scale real data further shows comparable performance, indicating that miniature-based data collection can serve as a viable alternative to collecting real training data from full-scale objects.

The contributions of this paper are summarized as follows:
\begin{itemize}

    \item We propose MiniPose, a pipeline that captures real RGB-D data from diverse poses using a scaled-down miniature replica, avoiding the physical difficulty of repeatedly repositioning large industrial objects.

    \item Extensive experiments and ablations show that MiniPose improves 6D pose estimation over synthetic-only training and outperforms depth noise modeling and correction.
    Both real RGB and real depth contribute to this gain, and MiniPose can serve as a practical alternative to full-scale real data.

\end{itemize}

The rest of this paper is organized as follows.
\Cref{sec:related} reviews related work, and \Cref{sec:method} introduces MiniPose for miniature-based data collection.
\Cref{sec:setup,sec:results} describe the experimental setup and results, respectively.
Finally, \Cref{sec:conclusion} concludes the paper and discusses future directions.

\section{Related Work}
\label{sec:related}

\subsection{6D Object Pose Estimation}
DenseFusion~\cite{wang2019densefusion} introduced dense pixel-wise RGB-D fusion with iterative pose refinement, while PVN3D~\cite{he2020pvn3d} estimates sparse 3D keypoints through point-wise offset voting.
FFB6D~\cite{he2021ffb6d} further develops bidirectional RGB--point-cloud fusion, and DFTr~\cite{zhou2023dftr} combines Transformer-based cross-modal fusion with weighted vector-wise keypoint voting.
In contrast to these sparse keypoint-voting approaches, RDPN6D~\cite{hong2024rdpn6d} regresses dense object coordinates for visible pixels using a residual representation.
These methods have been validated primarily on small- to medium-sized benchmarks such as YCB-Video~\cite{xiang2018posecnn} and LM-O~\cite{brachmann2014lmo}.
Although recent studies have begun to explore 6D pose estimation in industrial settings~\cite{Kalra_2024_CVPR,Balasubramanian_2025_Sensors}, physically large and heavy industrial components remain comparatively underrepresented.
We evaluate the proposed pipeline with two pose estimators, PVN3D, which uses keypoint voting, and RDPN6D, which uses dense correspondence regression, to examine the effect of miniature data across both approaches.

\subsection{Miniature-Based Approaches}
Miniature-based experimental platforms have been actively utilized, particularly in the autonomous driving domain.
MiniCity~\cite{buckman2022minicityp} is a 1:10-scale urban platform built to evaluate perception algorithms, including object detection and localization.
One study~\cite{jang2019simulation} transferred a simulation-trained driving policy to a 1:25-scale physical environment in a zero-shot manner, and another~\cite{tiedemann2022miniature} reported that miniature platforms can serve as a useful intermediate stage between simulation and real-world testing.
However, these works focus on executing or validating algorithms within the miniature environment itself.
In contrast, we use the miniature not as a validation environment but as a data collection tool, aiming at applications in the full-scale real environment.
Using miniatures as a source of real sensor data for the full-scale domain remains largely unexplored.
This work applies it to pose estimation of large industrial objects.

\subsection{Depth Noise Modeling and Correction}
\label{sec:related_depth}
Depth noise contributes to the sim-to-real gap because real sensors produce measurement errors that clean synthetic depth does not have.
Prior work has addressed this noise mainly in two ways, either by explicitly modeling it or by correcting the real measurements themselves.
Explicit noise modeling aims to reproduce measurement errors that vary with scene properties such as surface geometry and reflectance~\cite{landau2016simulating,meister2013}, but accurately reproducing these sensor-specific effects remains challenging~\cite{zhang2023active}.
The second strategy directly denoises the acquired depth measurements.
A learning-based method has been proposed to remove multipath distortions from ToF depth images in a robotic arm setup~\cite{son2016learning}, and the CDM has been trained on noisy-clean depth pairs generated through camera-noise modeling to denoise raw camera measurements~\cite{liu2026manipulation}.
However, such learned correction models are inherently tied to the sensors and noise distributions represented during training.
In contrast, our approach captures a miniature directly with a real RGB-D sensor, introducing real observation characteristics into training without a noise model, and geometrically transfers its depth to the full-scale domain.
We compare against both strategies, noise modeling and measurement correction, in~\Cref{sec:main_results}.

\section{Method}
\label{sec:method}

MiniPose collects real RGB-D data from a scaled-down replica of the target object and converts it into training data for the full-scale object, as illustrated in~\Cref{fig:pipeline}.
The replica is made from the same CAD model and paint as the full-scale object and is captured from diverse poses with a real RGB-D camera at a proportionally shorter distance.
Because the object and the camera distance are scaled together, the RGB images can be used as they are, and the depth maps are rescaled by the inverse of the miniature scale to match the full-scale range.
The converted data are then combined with synthetic data rendered from the full-scale CAD model (\Cref{sec:dataset}) to train two 6D pose estimators, PVN3D and RDPN6D.

\subsection{Miniature-Based Data Collection}
\label{sec:miniature}

Because a miniature replica can be freely repositioned, we can capture it with a real RGB-D sensor across a broad range of object poses.
Using the same CAD model as the full-scale object, we fabricate a scaled-down replica and coat it with the same paint as the full-scale pipe to approximately match surface color and reflectance.
The miniature is placed on the floor and captured from diverse viewpoints with a real RGB-D camera, while its position and orientation are randomly varied between captures.

\label{sec:depth_scaling}
The miniature observations are mapped to the full-scale domain using the scale relationship under a calibrated pinhole camera model, with lens distortion corrected before depth processing. Let $s\in(0,1)$ denote the miniature scale. Under proportional camera placement, scaling both the object and the camera--object distance by the same factor preserves the perspective relationship of corresponding points in the image. As a result, the miniature RGB images can be used without additional geometric scaling.

For depth, when the object and camera distance are both scaled by $s$, the measured values scale proportionally and are transformed to the full-scale domain using the inverse scale factor $1/s$. For example, with $s=0.5$, the depth values are scaled by a factor of 2. Under the ideal pinhole model, this transformation maps the miniature depth to the corresponding full-scale geometry. This geometric correspondence does not imply identical sensor measurement characteristics, so the rescaled depth is treated as an approximation of full-scale observations.

When depth scaling is applied, background depth values and their measurement variations are also scaled by $1/s$.
To reduce their influence on training, we mask the pipe object using SAM 2~\cite{ravi2024sam2}, crop a bounding box with a 20\,px margin, and use only the cropped region for training.

\begin{figure*}[t]
    \centering
    \includegraphics[width=0.95\textwidth]{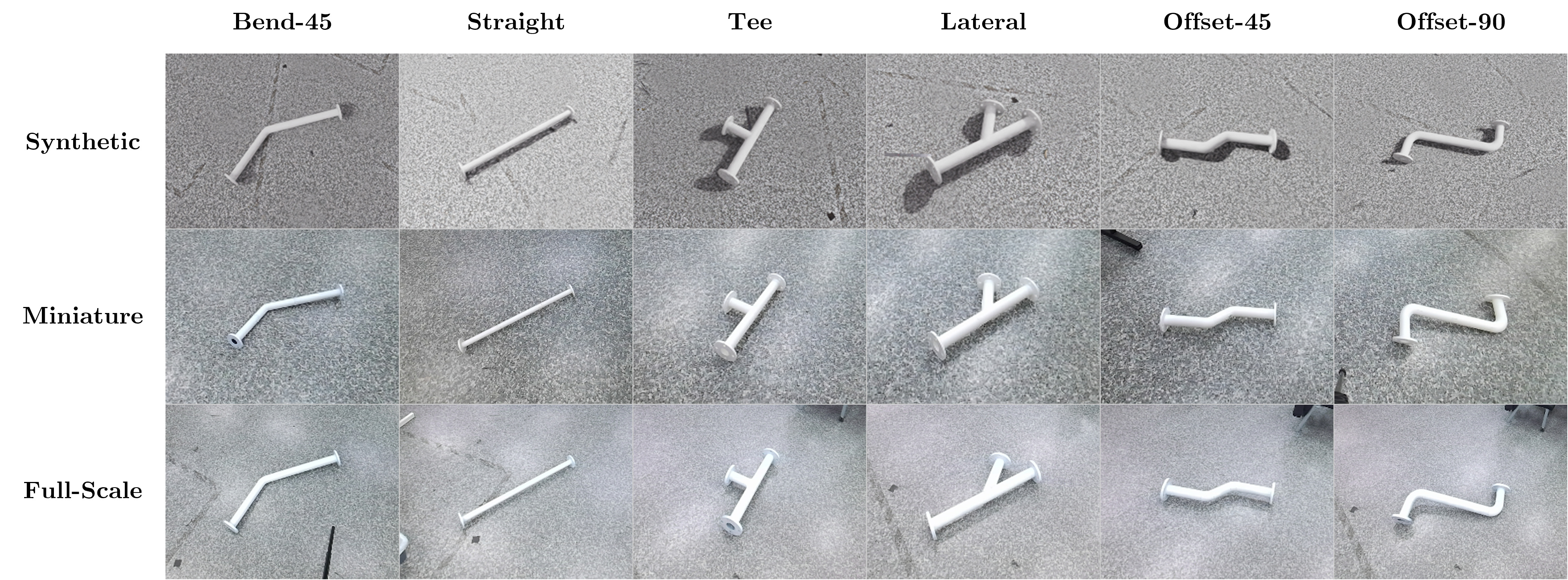}
    \caption{\textbf{Pipe geometries across data domains.} Representative examples of each pipe type in the synthetic, miniature, and full-scale domains (top to bottom) for Bend-45, Straight, Tee, Lateral, Offset-45, and Offset-90 (left to right).}
    \label{fig:pipe_types}
\end{figure*}

\subsection{Pose Estimators and Symmetry Handling}
\label{sec:symmetry_training}

We use PVN3D~\cite{he2020pvn3d} and RDPN6D~\cite{hong2024rdpn6d} as the 6D pose estimators, both with the SAM 2--cropped input described in~\Cref{sec:depth_scaling}.
We use the official RDPN6D implementation and enable its built-in symmetry-aware option for the symmetric classes (Straight, Tee, Offset-45, and Offset-90), following the official symmetric-object setting.

The original PVN3D samples 12,288 points from the full $640 \times 480$ RGB-D image for input to PointNet++~\cite{qi2017pointnetpp}.
The SAM 2 masking and cropping described in~\Cref{sec:depth_scaling} substantially reduce the pool of valid pixels.
Sampling 12,288 points from this reduced pool would result in a high proportion of duplicate points, degrading point diversity.
We therefore reduce the PVN3D PointNet++ input to 2,048 points.

The original PVN3D has no corresponding symmetry handling, so we modify its keypoint loss as follows.
For the symmetry-aware formulation, let $p_i$ denote the $i$-th of $N$ visible object points, let $M=8$ be the number of keypoints, and let $o_i^{\,j}$ denote the predicted translation offset from $p_i$ toward keypoint $j$.
Let $o_i^{\,j*}=k_j-p_i$ be the target offset toward the ground-truth keypoint $k_j$.

\begin{samepage}
The keypoint loss compares the predicted offset with the target offset over the visible object points as
\begingroup
\setlength{\abovedisplayskip}{4pt}
\setlength{\abovedisplayshortskip}{4pt}
\setlength{\belowdisplayskip}{4pt}
\setlength{\belowdisplayshortskip}{4pt}
\begin{equation}
    \mathcal{L}_{keypoints}=\frac{1}{N}\sum_{i=1}^{N}\sum_{j=1}^{M}
    \left\|o_i^{\,j}-o_i^{\,j*}\right\|_1
    \label{eq:kploss}
\end{equation}
\endgroup
\end{samepage}

The Straight, Tee, Offset-45, and Offset-90 pipes are invariant to symmetry transformations such as a $180^\circ$ rotation, so a single observation can correspond to multiple symmetrically equivalent poses.
If only one pose is used as the ground truth, the model may be unnecessarily penalized during training even when it predicts a symmetrically equivalent pose.
Let ${}^{C}k_j$ denote the ground-truth keypoint in the camera frame, ${}^{C}\mathbf{T}_{O}$ the $4\times4$ ground-truth transformation from object frame $O$ to camera frame $C$, and ${}^{O}\mathbf{T}_{C}$ the transformation in the reverse direction.
The transform ${}^{O}\mathbf{G}^{(m)}$ denotes the $m$-th symmetry transformation of the CAD model in the object frame, with $m$ indexing the symmetry-equivalent poses.
\begin{equation}
\begin{aligned}
    {}^{C}k_j^{(m)}
    &= {}^{C}\mathbf{T}_{O}\,{}^{O}\mathbf{G}^{(m)}\,{}^{O}\mathbf{T}_{C}\,{}^{C}k_j, \\
    o_i^{\,j*(m)}
    &= {}^{C}k_j^{(m)}-p_i
\end{aligned}
\label{eq:symcand}
\end{equation}
Equation~\eqref{eq:symcand} first maps ${}^{C}k_j$ to the object frame with ${}^{O}\mathbf{T}_{C}$, applies the symmetry transformation ${}^{O}\mathbf{G}^{(m)}$, and maps it back to the camera frame with ${}^{C}\mathbf{T}_{O}$.
Each symmetry transformation preserves the model point set, so the resulting candidates are equivalent with respect to the observed object geometry.

\begin{samepage}
The symmetry-aware keypoint loss is then defined as
\begin{equation}
    \mathcal{L}_{keypoints}^{\mathrm{sym}}
    = \min_{m}\frac{1}{N}\sum_{i=1}^{N}\sum_{j=1}^{M}
    \left\|o_i^{\,j}-o_i^{\,j*(m)}\right\|_1
    \label{eq:symloss}
\end{equation}
\end{samepage}
The symmetry-equivalent target closest to the prediction is used for supervision, while the non-symmetric classes (Bend-45 and Lateral) retain the original single ground-truth target.
All remaining loss terms and training hyperparameters follow the original PVN3D~\cite{he2020pvn3d} configuration.

\begin{figure*}[t]
    \centering
    \includegraphics[width=\textwidth]{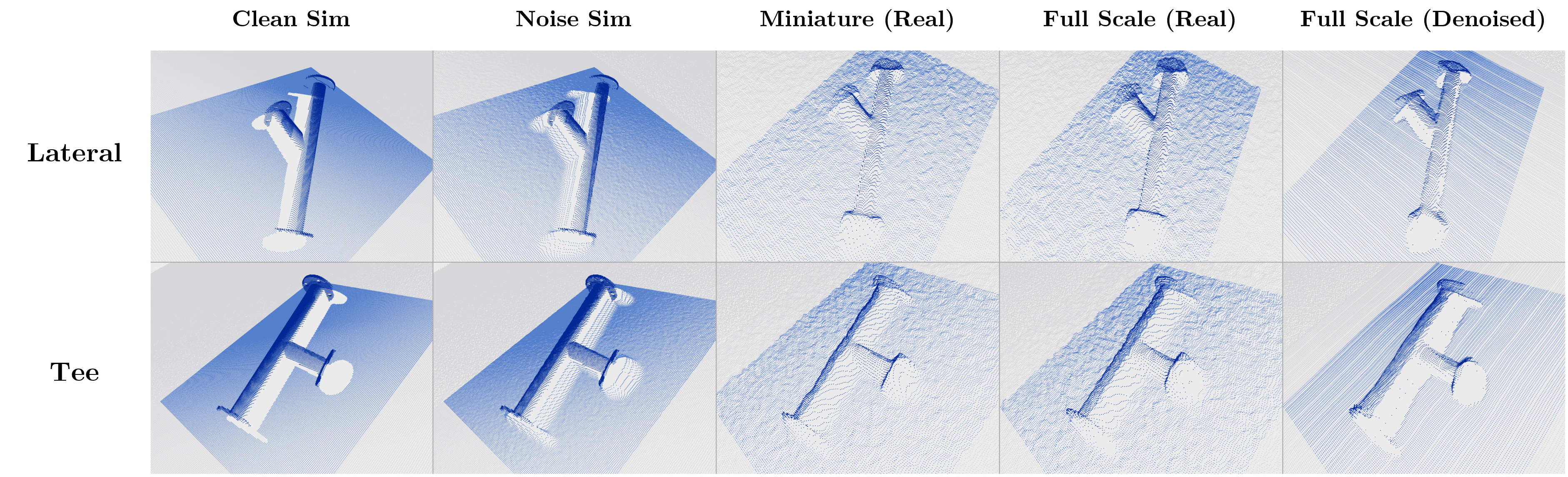}
    \caption{\textbf{Point cloud comparison across data types.} Clean Sim, Noise Sim (synthetic data with the ToF noise model), Miniature (Real), Full Scale (Real), and Full Scale (Denoised) by the CDM, shown for Lateral and Tee.}
    \label{fig:tof_noise}
\end{figure*}

\section{Experimental Setup}
\label{sec:setup}

\subsection{Dataset}
\label{sec:dataset}
We evaluate our method on six pipe types representative of the major structural forms found in industrial settings (Bend-45, Straight, Offset-45, Offset-90, Tee, and Lateral).
The types include a single bend (Bend-45), no bend (Straight), offset double bends (Offset-45 and Offset-90), and branching geometries (Tee and Lateral).
Representative examples of each pipe type across the three data domains used in this study are shown in~\Cref{fig:pipe_types}.
The pipes used in this study measure up to $1.5\,\text{m} \times 1.5\,\text{m} \times 0.5\,\text{m}$ and weigh approximately 10\,kg on average, far larger and heavier than the objects in common benchmarks such as YCB-Video~\cite{xiang2018posecnn} and LM-O~\cite{brachmann2014lmo}.

\textbf{Synthetic data.} Synthetic data were generated from the full-scale CAD model using NVIDIA Isaac Sim~\cite{isaacsim}, with camera distances randomly sampled between 1.8 and 2\,m.
The lighting intensity and position and the pipe surface color were randomized for each rendering.
Object poses were obtained by dropping the pipe from the air with a randomized initial orientation and capturing it once it settled under simulated physics.
Instead of material randomization, the ground plane was textured with one of several real floor photographs from the data-collection site to reduce the RGB appearance gap relative to the test environment.
Each rendered frame yields an RGB image, a depth map, and the corresponding ground-truth 6D pose, for a total of 10,000 images per pipe type.

\textbf{Miniature data.} Miniature RGB-D data were captured with a ToF camera (Microsoft Azure Kinect) at distances of 0.9--1\,m using a 50\%-scale replica (about 100--200\,g) fabricated with a Bambu X1C FDM 3D printer, and then processed using the depth scaling described in~\Cref{sec:depth_scaling}.
This yielded 1,000 images per pipe type.
The Straight pipe is an exception, with 500 images, because its appearance changes little under rotation.
Invalid depth pixels (holes) in the captured depth maps were filled with the nearest valid depth.

\textbf{Full-scale real data.} Full-scale real data were captured at distances of 1.8--2\,m in a single environment, with 100 evaluation images per pipe type.
A disjoint set of the same size serves as the full-scale real training reference in the ablation (\Cref{sec:ablation}).

\textbf{Training set.} For training, the 10,000 synthetic images per pipe type are combined with the miniature images, yielding 11,000 training images per pipe type (10,500 for Straight).
The miniature subset therefore constitutes less than 10\% of the training data for each pipe type.
While the synthetic data provide broad coverage of the pose space, the miniature data complement them with real RGB-D observations from the target sensor.

\subsection{Evaluation Metric}
The pipe geometries evaluated in this study include rotationally symmetric structures, which introduce pose ambiguity along the axis of symmetry.
We therefore use ADD-S~\cite{xiang2018posecnn} as the evaluation metric, defined as
\begin{equation}
    \text{ADD-S} = \frac{1}{|\mathcal{M}|} \sum_{x \in \mathcal{M}} \min_{x' \in \mathcal{M}} \left\| (\hat{T} x) - (T x') \right\|
    \label{eq:adds}
\end{equation}
where $\mathcal{M}$ is the set of object surface points, $\hat{T}$ is the predicted pose, and $T$ is the ground-truth pose.
We report the ADD-S area under the curve (AUC), computed by integrating the fraction of model points whose ADD-S distances fall below each threshold up to 10\% of the object diameter.

\subsection{Baselines and Training Settings}
\label{sec:baselines}
The baselines are designed to examine how much depth noise affects training, and to show that mixing in miniature data helps even without explicit analytical modeling of depth noise.
They follow the two strategies reviewed in~\Cref{sec:related_depth}, namely test-time correction with CDM and noise modeling with a ToF noise model.
We compare four train--test settings for each of PVN3D and RDPN6D, with the corresponding data types shown in~\Cref{fig:tof_noise}.

\textbf{Setting 1.} Trains on 10,000 clean synthetic images per pipe type and evaluates on full-scale real data.

\textbf{Setting 2.} Trains on the same 10,000 clean synthetic images as Setting 1 and evaluates on full-scale real data whose depth is denoised by the pretrained CDM~\cite{liu2026manipulation}, used without additional training or fine-tuning.

\textbf{Setting 3.} Trains on synthetic images augmented with a ToF noise model and evaluates on full-scale real data.
The noise model consists of distance- and incidence-angle-dependent shot noise with spatial correlation and low-signal dropout, with parameters set from the Azure Kinect DK specification~\cite{azurekinectspec} and prior ToF literature~\cite{foix2011tof,lange2001solidstate}.
Flying pixels are modeled as a smooth depth transition on the background side of object boundaries.
The width and offset of this transition are fitted to the edge profile measured on the real miniature captures, not on the test sets.
Multipath and systematic (wiggling) errors are not modeled, so this setting is an approximate sensor-noise model rather than a complete reproduction of the real camera.

\textbf{Ours.} The proposed configuration trains on the clean synthetic images combined with 1,000 miniature RGB-D images per pipe type (500 for Straight) and evaluates on full-scale real data, with miniature images making up approximately 10\% of each batch for both estimators.

All PVN3D variants were trained on an RTX 5090 GPU with batch size 10 for 60 epochs, and all RDPN6D variants with batch size 24 for 80 epochs per class, with the same number of epochs across all settings of each model.

\begin{table}[t]
\centering
\footnotesize
\caption{ADD-S AUC on real pipes for two pose estimators under four training settings.}
\label{tab:main}
\setlength{\tabcolsep}{3.4pt}
\renewcommand{\arraystretch}{1.05}
\setlength{\aboverulesep}{0.15ex}
\setlength{\belowrulesep}{0.25ex}
\sbox{\tabbox}{\begin{tabular}{lcccccccc}
\toprule
Method & \multicolumn{4}{c}{PVN3D~\cite{he2020pvn3d}} & \multicolumn{4}{c}{RDPN6D~\cite{hong2024rdpn6d}} \\
\cmidrule(lr){2-5}\cmidrule(lr){6-9}
Setting & Set 1 & Set 2 & Set 3 & Ours & Set 1 & Set 2 & Set 3 & Ours \\
\midrule
Straight & 89.0 & 86.8 & \underline{91.9} & \textbf{94.5} & 94.0 & 93.8 & \textbf{94.7} & \underline{94.6} \\
Bend-45 & \underline{93.5} & 92.8 & 91.5 & \textbf{94.3} & \underline{95.2} & 94.5 & 95.1 & \textbf{95.5} \\
Tee & 92.3 & 89.8 & \underline{92.5} & \textbf{94.4} & 93.4 & 92.7 & \underline{93.5} & \textbf{94.9} \\
Lateral & 92.4 & 89.4 & \underline{92.6} & \textbf{94.8} & 92.6 & 92.1 & \underline{93.4} & \textbf{94.0} \\
Offset-45 & \underline{92.0} & 88.7 & 91.6 & \textbf{94.4} & 91.3 & 88.8 & \underline{92.5} & \textbf{93.5} \\
Offset-90 & 92.0 & 88.3 & \underline{92.2} & \textbf{94.8} & 94.2 & 91.5 & \underline{94.4} & \textbf{94.7} \\
\midrule
Avg (6) $\uparrow$ & 91.9 & 89.3 & \underline{92.1} & \textbf{94.5} & 93.5 & 92.2 & \underline{93.9} & \textbf{94.5} \\
\bottomrule
\end{tabular}}
\usebox{\tabbox}
\par\vspace{2pt}
\parbox{\wd\tabbox}{\scriptsize\raggedright Bold and underlined values indicate the best and second-best results within each estimator.}
\vspace{-6pt}
\end{table}

\begin{figure*}[t]
    \centering
    \includegraphics[width=\textwidth]{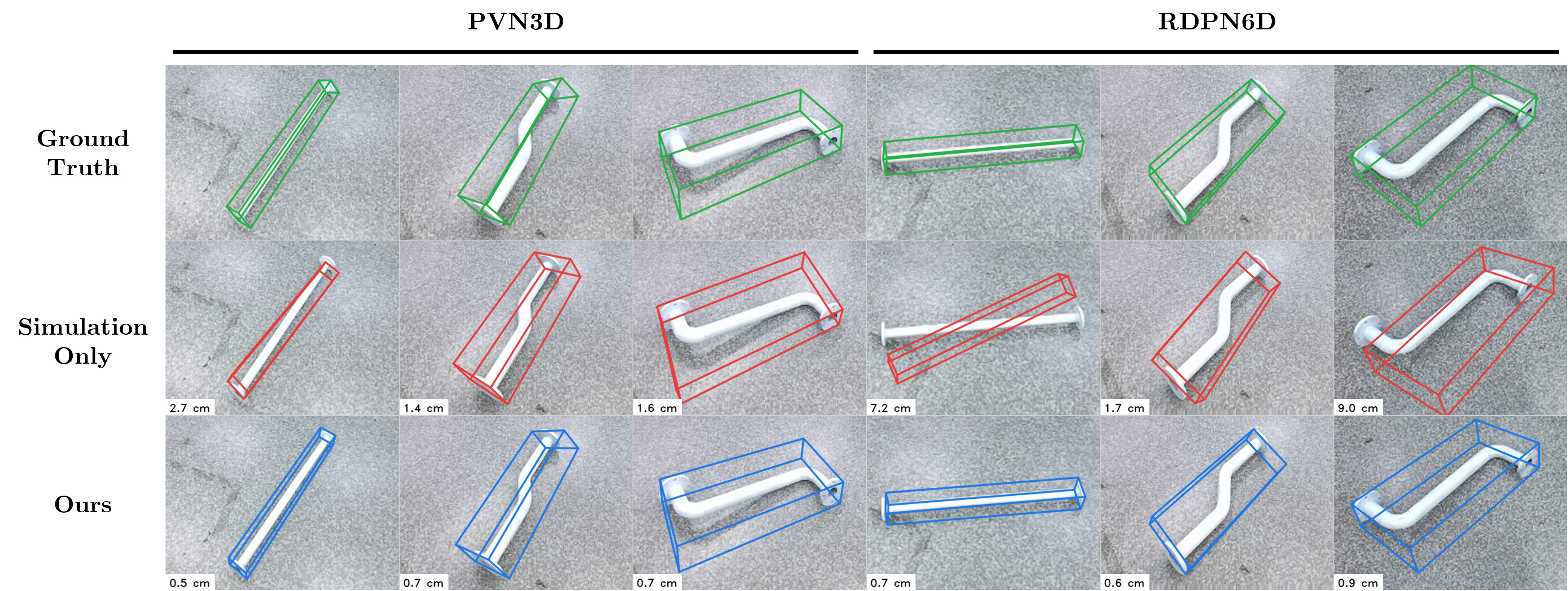}
    \caption{\textbf{Qualitative 6D pose estimation results on real pipes.} Green boxes denote the ground truth, while red and blue boxes show the inference results of the model trained on simulation data only (Simulation Only) and on simulation and miniature data (Ours), respectively. The left three columns show PVN3D and the right three show RDPN6D. The number in each image is the ADD-S error of that frame's inference result.}
    \label{fig:qualitative}
\end{figure*}

\begin{table*}[t]
\centering
\footnotesize
\caption{Ablation on synthetic depth noise, input modality, and real data.}
\label{tab:ablation}
\setlength{\tabcolsep}{9.8pt}
\renewcommand{\arraystretch}{1.05}
\setlength{\aboverulesep}{0.15ex}
\setlength{\belowrulesep}{0.25ex}
\sbox{\tabbox}{\begin{tabular}{lcccccccccc}
\toprule
Method & \multicolumn{5}{c}{PVN3D~\cite{he2020pvn3d}} & \multicolumn{5}{c}{RDPN6D~\cite{hong2024rdpn6d}} \\
\cmidrule(lr){2-6}\cmidrule(lr){7-11}
Setting & Set 4 & Set 5 & Set 6 & Set 7 & Ours & Set 4 & Set 5 & Set 6 & Set 7 & Ours \\
\midrule
Straight & 93.6 & 93.4 & 91.3 & \textbf{96.1} & \underline{94.5} & 94.6 & \underline{94.8} & \underline{94.8} & \textbf{96.2} & 94.6 \\
Bend-45 & 92.4 & 93.9 & \underline{94.1} & 93.5 & \textbf{94.3} & 95.0 & 95.5 & \underline{95.7} & \textbf{96.3} & 95.5 \\
Tee & 92.6 & 92.7 & \underline{93.4} & 92.5 & \textbf{94.4} & 93.4 & 93.2 & 94.3 & \textbf{95.2} & \underline{94.9} \\
Lateral & 88.1 & \underline{92.9} & 91.9 & 92.3 & \textbf{94.8} & 92.5 & 93.6 & 93.7 & \textbf{95.6} & \underline{94.0} \\
Offset-45 & 91.9 & \underline{93.5} & 93.0 & 93.0 & \textbf{94.4} & 91.4 & 92.6 & 91.2 & \textbf{94.4} & \underline{93.5} \\
Offset-90 & 89.4 & \underline{94.5} & 91.7 & 92.9 & \textbf{94.8} & 94.3 & 94.5 & 94.4 & \textbf{95.2} & \underline{94.7} \\
\midrule
Avg (6) $\uparrow$ & 91.3 & \underline{93.5} & 92.6 & 93.4 & \textbf{94.5} & 93.5 & 94.0 & 94.0 & \textbf{95.5} & \underline{94.5} \\
\bottomrule
\end{tabular}}
\usebox{\tabbox}
\par\vspace{2pt}
\parbox{\wd\tabbox}{\scriptsize\raggedright Bold and underlined values indicate the best and second-best results within each estimator.}
\end{table*}

\section{Results}
\label{sec:results}

\subsection{Main Results}
\label{sec:main_results}

We evaluate the four settings for both PVN3D and RDPN6D on each of the six pipe types and report the ADD-S AUC in~\Cref{tab:main}.
Setting 2 (test-time denoising with CDM) is simple to apply but does not improve performance, lowering the average AUC from 91.9 to 89.3 for PVN3D and from 93.5 to 92.2 for RDPN6D, which suggests that it may not be directly applicable to every environment.
Setting 3 (ToF noise model) improves over Setting 1 (clean simulation data).
It achieves the second-highest average AUC for both estimators (92.1 versus 91.9 for PVN3D and 93.9 versus 93.5 for RDPN6D), so analytical camera modeling appears to help training.

Ours (clean synthetic data with miniature RGB-D data) yields the largest improvement.
For PVN3D, the proposed setting achieves the highest AUC on every pipe type, raising the average to 94.5, with the largest gain on Straight (89.0 to 94.5).
For RDPN6D, the average also improves from 93.5 to 94.5, with gains on all six pipe types.
Setting 3 must be constructed through explicit mathematical modeling of the sensor and remains imperfect, whereas simply mixing miniature data into clean simulation data performs well without any camera-specific noise modeling.
Qualitative results in~\Cref{fig:qualitative} show that, in the examples shown, the predicted boxes with the proposed setting align more closely with the pipes than those of synthetic-only training.

\subsection{Ablation Study}
\label{sec:ablation}

To examine which properties of the miniature data improve real-world performance, we compare four additional training settings in~\Cref{tab:ablation} against Ours.

\textbf{Setting 4.} Adds Gaussian noise ($\sigma = 2.5$\,mm) to synthetic depth to test whether simple artificial noise can substitute for real sensor data.

\textbf{Settings 5 and 6.} These settings separate the contributions of real RGB and real depth in the miniature data.
For each miniature frame, we render the CAD model with the same camera parameters and object pose as the captured frame.
Setting 5 keeps the captured RGB and replaces the depth map with the rendered depth, whereas Setting 6 keeps the captured depth and replaces the RGB image with the rendered RGB.

\textbf{Setting 7.} Adds 100 directly captured full-scale real images per pipe type and serves as a reference for conventional sim-to-real training.
As with Ours, these images make up approximately 10\% of each batch.

On average, Setting 4 (Gaussian noise) does not improve over Setting 1 (clean-synthetic training) in~\Cref{tab:main} (91.3 vs 91.9 for PVN3D and 93.5 vs 93.5 for RDPN6D), so the characteristics of real depth do not appear to be reproduced by simple independent noise.
In contrast, Settings 5 and 6 (real RGB only and real depth only) both improve the average AUC.
For PVN3D, the average is 93.5 with Setting 5 and 92.6 with Setting 6, while both achieve 94.0 for RDPN6D.
Using both modalities (Ours) gives the highest average among the settings without full-scale real data (94.5 for both estimators).
Setting 7 (full-scale real data) yields a lower average than Ours for PVN3D (93.4 vs 94.5), whereas it remains higher for RDPN6D (95.5).
Even so, obtaining the 100 full-scale images per pipe type already required repositioning large pipes within a large workspace, whereas the miniature data could be captured in a small space.
The miniature data therefore offer a promising option where collecting data from large objects is difficult.

\section{Conclusion}
\label{sec:conclusion}

We presented MiniPose, a practical pipeline for acquiring real RGB-D training data for 6D pose estimation of large industrial objects. MiniPose captures an easily handled scaled-down replica with a real RGB-D camera and maps the miniature depth to full-scale geometry through depth rescaling.
The key advantage of this approach is that it retains the broad pose coverage provided by synthetic data while making it easier to incorporate real appearance and sensor measurement characteristics into training without requiring a noise model of the camera.
Experiments on six industrial pipe types with two pose estimators, PVN3D and RDPN6D, show that incorporating miniature data improves the average ADD-S AUC over synthetic-only training for both.
The proposed approach achieves comparable performance to training with directly captured full-scale real data, although the relative performance varies by estimator, suggesting that miniature-based data collection can reduce the burden of acquiring real training data for large objects.

Several directions remain for future work.
First, all experiments in this study use a single ToF camera (Microsoft Azure Kinect), and whether the proposed pipeline generalizes to other depth sensors remains to be validated.
Second, because sensor noise does not scale linearly with distance, the rescaled miniature depth only approximates full-scale measurements, and how this difference affects the learned model remains to be examined.
Third, the miniature scale was tested only at 0.5, and evaluation across a wider range of scales is needed.




\bibliographystyle{IEEEtran}
\bibliography{references}

\end{document}